\documentclass[sigconf]{acmart}

\AtBeginDocument{%
  \providecommand\BibTeX{{%
    \normalfont B\kern-0.5em{\scshape i\kern-0.25em b}\kern-0.8em\TeX}}}

\setcopyright{licensedothergov}
\copyrightyear{2022}
\acmYear{2022}
\acmDOI{10.1145/3485447.3511941}

\acmConference[WWW '22]{Proceedings of the ACM Web Conference 2022}{April 25--29, 2022}{Virtual Event, Lyon, France}
\acmBooktitle{Proceedings of the ACM Web Conference 2022 (WWW '22), April 25--29, 2022, Virtual Event, Lyon, France}
\acmPrice{15.00}
\acmISBN{978-1-4503-9096-5/22/04}

\usepackage{algorithm}
\usepackage{algorithmic}

\usepackage{microtype}

\usepackage{amsmath}

\usepackage{color}
\usepackage{multirow}
\usepackage{textgreek}
\usepackage{graphics}
\usepackage{pifont}
\graphicspath{ {./} }

\newcommand\BibTeX{B\textsc{ib}\TeX}

\newcommand{\norm}[1]{\left\lVert#1\right\rVert}

\usepackage{fixltx2e}
\usepackage[export]{adjustbox}
\usepackage{caption}
\usepackage{subcaption}
\usepackage{tikz}
\newcommand\mybox[2][]{\tikz[overlay]\node[fill=blue!20,inner sep=2pt, anchor=text, rectangle, rounded corners=1mm,#1] {#2};\phantom{#2}}

\begin{document}

\title{AR-BERT: Aspect-relation enhanced Aspect-level Sentiment Classification with Multi-modal Explanations}

\author{Sk Mainul Islam}
\affiliation{%
  \institution{IIT Kharagpur \country{India}}
  \postcode{721302}
}

\author{Sourangshu Bhattacharya}
\affiliation{%
  \institution{IIT Kharagpur \country{India}}
  \postcode{721302}
}


\begin{abstract}
Aspect level sentiment classification (ALSC) is a difficult problem with state-of-the-art models showing less than $80\%$ macro-F1 score on benchmark datasets.
Existing models do not incorporate information on aspect-aspect relations in knowledge graphs (KGs), e.g. DBpedia. Two main challenges stem from inaccurate disambiguation of aspects to KG entities, and the inability to learn aspect representations from the large KGs in joint training with ALSC models.
 We propose AR-BERT, a novel two-level global-local entity embedding scheme that allows efficient joint training of KG-based aspect embeddings and ALSC models.
A novel incorrect disambiguation detection technique addresses the problem of inaccuracy in aspect disambiguation.
We also introduce the problem of determining mode significance in multi-modal explanation generation, and propose a two step solution.
The proposed methods show a consistent improvement of $2.5 - 4.1$ percentage points, over the recent BERT-based baselines on benchmark datasets. The code is available at \url{https://github.com/mainuliitkgp/AR-BERT.git}.
\end{abstract}




\maketitle

\section{Introduction}

\begin{figure*}
\centering
    {\small
    \includegraphics[width=16cm]{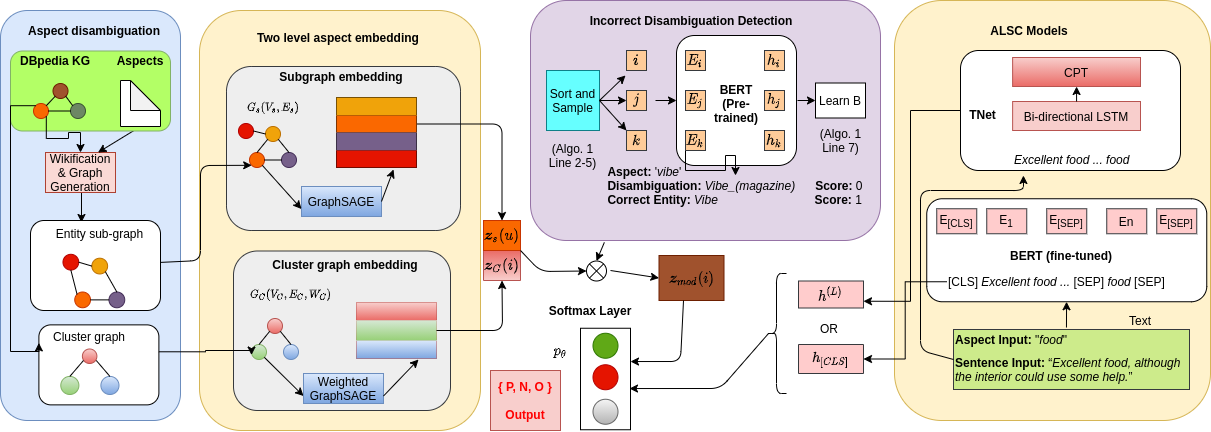}
    \caption{Architecture of our framework for end-to-end training of ALSC model while incorporating aspect relations from large KGs. Yellow backgound denotes modules trained end-to-end, green backgound denotes input.}
    \label{Figure:2}
    }
\end{figure*}

Aspect level sentiment classification (ALSC) is an important NLP task \citep{DBLP:conf/kdd/HuL04, DBLP:conf/semeval/PontikiGPPAM14, DBLP:conf/acl/DongWTTZX14}, where we predict the sentiment portrayed in a \textit{sentence} (also called context) towards an identified \textit{aspect} phrase. 
Recently, models capturing aspect-specific features, e.g., Transformation Network (TNet) \citep{DBLP:conf/acl/LamLSB18}, which constructs aspect-specific embedding of context words, or BERT-based models  \citep{DBLP:conf/naacl/DevlinCLT19}, which capture aspect-specific representations of sentences, have outperformed previous sequential prediction models.
Other recent improvements include domain adaptation of the BERT model \citep{DBLP:conf/lrec/RietzlerSOE20} and incorporating entity relationships within the same sentence using graph convolutional networks \citep{DBLP:journals/kbs/ZhaoHW20}. However, existing ALSC methods do not explicitly utilize the relations between aspects, which could potentially lead to better performance.


We observe that many of the aspect phrases, e.g., \texttt{Windows 8, Mozzarella, Taylor Swift}, etc., are mentions of named entities appearing in knowledge graphs (KG), e.g., DBpedia, which encode various entity-entity relations.
While some of the aspects may be unseen in the training data, their neighbors (related aspects) in the KG may be abundant. This information can be used to infer important signals from the context sentence, which in turn can help in the prediction of correct polarity. 
For example, in the sentence, \texttt{My laptop with Windows 7 crashed and I did not want Windows 8.}, the aspect \texttt{Windows 7} has only 17 examples in the training data. The current state-of-the-art ALSC model \citep{DBLP:conf/lrec/RietzlerSOE20} wrongly predicts the aspect sentiment as positive. However, its related aspects (1-hop neighbors in the DBpedia KG) have 209 training examples, which can lead to the correct prediction of sentiment.
Our primary objective is to incorporate the KG relations between aspects into ALSC models. 

The main challenges in model design are: (1) end-to-end training of ALSC models with KG embeddings is infeasible due to large scale of KGs, and (2) most scalable off-the-shelve named entity disambiguation techniques, e.g., wikifier \citep{DBLP:journals/informaticaSI/BrankLG18} are highly inaccurate.
While state of the art named entity disambiguation methods \citep{DBLP:conf/aaai/KarRBDC18, DBLP:conf/conll/Broscheit19} are  accurate, they still do not scale to the entire DBpedia KG. 
We solve the problem of learning aspect representations from large KGs using a two-level graph embedding technique: one corresponding to a higher level \textit{cluster graph}, and another for subgraphs. These embeddings can be efficiently trained jointly along with ALSC models.
The problem of inaccurate wikification \citep{DBLP:journals/informaticaSI/BrankLG18} method for aspect disambiguation, is ameliorated by a novel probing function based detection of incorrect aspect disambiguations.
Figure \ref{Figure:2} shows the overall architecture of the proposed technique. 

The deep multi-modal ALSC model proposed here utilizes information from both text and graph data in an opaque manner, thus reducing the trust in the model predictions. Hence, we seek to design a postthoc global explanation model for predicting multi-modal explanations from both context words, and KG-entities.
A key challenge in the prediction of multi-modal explanations is the prediction of importance of the mode for which explanations are being generated, since there may not be any valid explanations from a given mode. We design an explanation model  which also predicts the mode importance, and a  two-step prodecure for jointly learning the unimodal explanation models, as well as the mode importance predictor.
To the best of our knowledge, ours is the first model for jointly predicting multi-modal explanations from text and graph data.

Experimental results show that proposed ALSC models improve the macro-F1 score and accuracy of state-of-the-art ALSC methods on three benchmark datasets by between $ 2.5\% - 4.1\%$. We also demonstrate that the scarcity of training examples is indeed a factor for the inaccuracy of existing models.
We also show that classification accuracy of wrongly disambiguated aspects improves significantly with the disambiguation correction method.
Finally, experimental results using the explanation prediction model show that both predicted explanations and significant mode are effective and intuitive.
To summarize, our main contributions are:
(1) AR-BERT - a scalable aspect-relation enhanced ALSC algorithm.
(2) A novel technique for detecting incorrect entity disambiguations.
(3) A multi-modal explanation generation model with significant mode detection.


\section{Related Work}

\noindent \textbf{ALSC with Graph Embedding:}
\citep{majumder2018iarm} uses memory networks generate aspect representations influenced by other aspects in the same sentence.
\citep{zhang2019aspect} uses aspect-specific GCN, and 
\citep{liang2020jointly} uses an ``interactive'' dependency graph to capture the relations between aspect in a sentence.
\citep{tang2020dependency} also encodes information in dependency graph using a transformer-like network.
However, none of the above methods can be used at a scale where we can apply it to a knowledge graph like DBpedia.
\citep{Xu20aspectsentiment},\citep{jiang2020metnet} focus on determining aspect specific opinion spans. 
In addition to the models described in section \ref{sec:background_alsc}, neural network models such as Memory Networks \citep{DBLP:conf/emnlp/TangQL16, DBLP:conf/acl/LiuCWMZ18, DBLP:conf/emnlp/ChenSBY17}, LSTM-based models \citep{DBLP:conf/emnlp/WangHZZ16, DBLP:conf/ijcai/MaLZW17, DBLP:conf/eacl/ZhangL17}, and 
Capsule Networks \citep{DBLP:conf/emnlp/DuSWQLXL19} have also been explored for ALSC.

\textbf{Knowledge graphs in BERT representations:} \citet{DBLP:conf/nips/ReifYWVCPK19} measures the word sense similarities using a semantic probe on word embeddings. \citet{DBLP:conf/naacl/HewittM19} shows that contextual word embedding incorporates syntactic informations.
\citet{DBLP:conf/conll/Broscheit19} investigates entity knowledge in BERT embedding. 
\citep{poerner2020bert, peters2019knowledge, zhang2019ernie} propose a promising line of schemes for incorporating entity knowledge in KGs into BERT embeddings. However end-to-end training with these methods has to take entire KG into account, and is expected to be computationally expensive.\\
\citep{sun2020colake} modifies the BERT encoder and training procedure in order to incorporate graphs constructed from KG and unstructured text. However, this is not scalable.
\citep{liu2020k} augments the unstructured text with triples from KG, and trains BERT on the resulting corpora. This technique is also not scalable when trying to capture relations between all entities.


\textbf{Explainable models:} 
For explaining textual and graph information in a unimodal manner popular methods e.g. LIME \citep{lime} (extracts important words for a particular prediction) and GNNExplainer \citep{ying2019gnnexplainer} (extracts an important subgraph for a particular node classification prediction), cannot be used since they only learn local explanation models. We use the recently proposed methods of concept learning \citep{yeh2019completeness} for explanations using textual data, and PGExplainer \citep{luo2020parameterized} for explanations using graphs data.
While multi-modal explanation generation has been studied in for image and text modes \citep{park2018multimodal, Kanehira_2019_CVPR}, 
to the best of our knowledge, multi-modal explanation and mode significance has not been studied in the context of textual and graph data. 

\section{Explainable ALSC with Aspect Relation}



In this section, we describe our approach for improving the performance of aspect level sentiment classification (ALSC) methods using semantic relations between aspects which can be extracted from Knowledge Graphs (KG), e.g. DBpedia.
The key motivation behind our work is that certain aspects are not well represented by examples in the training set, but they have neighboring entities in the KG which have more examples. Hence, the semantic information learned from the neighboring aspect may be transferred to the current aspect through aspect embeddings. For example, in the sentence \texttt{[However, I can refute that OSX is "FAST"]}, the aspect \texttt{OSX} has the corresponding DBpedia entity \textit{MacOS}, which has only $7$ examples in the training set. However, \textit{MacOS} has a related entity \textit{Microsoft\_Windows} which has $37$ examples. This leads to the existing BERT-based ALSC method \cite{DBLP:conf/lrec/RietzlerSOE20} misclassifying this example as positive sentiment polarity based on the context word \textit{FAST}, 
whereas our method focusses on the context word \textit{refute} and classifies the example correctly as negative sentiment polarity.

Our method has 2 broad components: (1) disambiguating mentions of aspects (e.g. \textit{OSX}) to entities from a KG (e.g. DBpedia entity \textit{MacOS}), and representing them as an embedding vector,
and (2) incorporation of the vector representation of the aspects into state of the art ALSC models, e.g. TNET and BERT, using end-to-end training. 
Figure \ref{Figure:2} describes the overall architecture of our technique.
Section \ref{sec:background_alsc} provides the background on ALSC, sections \ref{sec:Target-Relation-Incorporation}, \ref{sec:end-to-end}, and \ref{sec:disambiguation-correction} describe the proposed ALSC model, AR-BERT, and an incorrect disambiguation detection technique.
Section \ref{sec:multimodal-explanation} describes a novel model for multi-modal explanation generation for AR-BERT.

\subsection{Background in ALSC} \label{sec:background_alsc}

\newcommand{\bD}{\mathbf{D}}
\newcommand{\bz}{\mathbf{z}}
\newcommand{\vw}{\Vec{w}}
\newcommand{\vx}{\Vec{x}}
\newcommand{\vh}{\Vec{h}}

The task of aspect level sentiment classification (ALSC) is to determine the sentiment polarity $y \in \{P, N, O\}$ of an input sentence $w$ for an aspect phrase $w^t$, which is a part of the input sentence.
Here, $P$, $N$, and $O$ correspond to positive negative and neutral sentiment respectively.
ALSC models take representations of the context $w$, $\vx=(x_1,...,x_n)$, and that of the aspect $w^t$, $\vx^t=(x_1^t,...,x_m^t)$ as inputs.
Most state-of-the-art ALSC models, including TNET \citep{DBLP:conf/acl/LamLSB18}, and BERT \citep{DBLP:conf/naacl/DevlinCLT19} transform the context representation using an aspect representation to finally arrive at an aspect-specific representation for context words.
Here, $n$ denotes the length of the (context) sentence and $m$ denotes the length of the aspect.
We briefly describe the architectures of these methods.

TNet  \citep{DBLP:conf/acl/LamLSB18} consists of three sequential modules (sets of layers):
The first module is a Bi-LSTM layer which takes context embeddings $\vx$ and aspect (target) word embeddings  $\vx^t$ corresponding to each eaxmple and outputs the contextualized representations 
$\vh^{(0)}= ( h_1^{(0)}(\vx),...,h_n^{(0)}(\vx) )$ 
and 
$\vh^t= ( h_1^t(\vx^t),...,h_m^t(\vx^t) )$ respectively where $h_i^{(0)}(x), h_j^t(x^t) \in \Bbb R^{2 D_h}$, $i \in \{1, \dots ,n\}$, $j \in \{1,\dots , m\}$.
The second module contains $L$ layers of context preserving transformations (CPT) blocks. In each layer $l$ the aspect representation is first transformed into aspect specific representation as $r_i^t= \sum_{j=1}^{m} h_j^t * \mbox{SoftMax}(h_i^{(l)}, h_j^t)$, 
then incorporated into context representation as $\tilde{h_i}^{(l)} = \mbox{FeedForward}(h_i^{(l)} r_i^t)$, 
and finally passed into context preserving \textbf{LF/AS} block to get the output of next layer:  $h_i^{(l+1)}=LF/AS(h_i^{(l)}, \tilde{h_i}^{(l)})$. 
The third module uses convolution and pooling layers on position-aware encodings to produce a fixed dimensional vector $z$.

BERT has been applied to ALSC by \citet{DBLP:conf/lrec/RietzlerSOE20} and \citet{DBLP:conf/naacl/SunHQ19}, where they 
model the sentiment classification task as a sequence-pair classification task.  The input sentence ($\vx$) and aspect phrase ($\vx^t$) are encoded as [CLS] $\vx$ [SEP] $\vx^t$ [SEP].
The last layer hidden representation of CLS token $h_{[CLS]} \in \Bbb R^{768}$, which is the aspect-aware representation of the input sequence, is used for the downstream classification task. The sentiment polarity distribution is predicted using a feedforward layer with softmax activation and trained using the cross-entropy loss.
Recently, SDGCN-BERT \citep{DBLP:journals/kbs/ZhaoHW20} has been proposed to capture sentiment dependencies between multiple aspects in a sentence using a graph convolution network. BERT-ADA \citep{DBLP:conf/lrec/RietzlerSOE20} uses BERT domain-specific language model fine-tuning for ALSC and results in best accuracy on some benchmark datasets. In this paper we build on TNet, BERT, SDGCN-BERT, and BERT-ADA, to incorporate knowledge from KG. Next, we describe our framework for the scalable incorporation of KG information in ALSC.

\subsection{Aspect Relation Incorporation from KG} \label{sec:Target-Relation-Incorporation}

Incorporating aspect relation from KG into ALSC models has two substeps: (1) Aspect to entity mapping and (2) Computation of entity embedding. 
The first step involves the identification of Wikipedia entities corresponding to an aspect word in a context. This problem is solved by named entity disambiguation (NED) or wikification. We use \textit{wikifier}
API \citep{DBLP:journals/informaticaSI/BrankLG18} for this purpose. Note that, here we use a freely available and computationally efficient method for entity linking, at the cost of accuracy. We partially make up for the loss of accuracy in the posthoc disambiguation correction described in section \ref{sec:disambiguation-correction}.

For learning the entity embeddings (step 2), we use the popular GraphSAGE algorithm
\citep{DBLP:conf/nips/HamiltonYL17}, which is applicable for both supervised and unsupervised tasks. 
The entity relation graph is generated using the DBpedia \footnote{https://wiki.dbpedia.org/downloads-2016-10} page links knowledge graph, where each vertex is an entity in the DBpedia KG, and an edge is a tuple of the form $<Sub, Pred, Obj>$ where $Sub$ and $Obj$ are the subject and object entities, and $Pred$ is the predicate relation between $Sub$ and $Obj$.
However, the whole DBpedia knowledge graph (KG) is too large (with $\sim$ 22 million nodes and $\sim$ 173 million edges) to embed using deep NRL techniques.
Another alternative is to consider the subgraph $G$ induced by entities present in the ALSC training dataset only. The problem with this subgraph is that it is disconnected. Hence, the similarity preserving embeddings of entities are only consistent within the connected components of $G$. This may lead to two very different entities $u$ and $v$ accidentally ending up close to each other.
In this section, we describe a two-level scalable network embedding technique that scales to DBpedia while avoiding the above-mentioned problems.

\subsubsection{Two-level Aspect Entity Embedding} \label{sec:two-level entity embedding}

The key idea behind two-level aspect embedding (representations) is two use two smaller graphs constructed from the large KG: (1) a cluster graph $G_C(V_C, E_C, W_C)$: which captures the global connectivity structure between clusters of entities, and (2) the subgraph $G_s(V_s, E_s)$  induced by aspects (entities) in the training dataset.
Note that since the subgraph $G_s$ can be disconnected, we need a combination of \textit{cluster graph embedding} $\bz_C(u)$, and \textit{subgraph  embedding}, $\bz_s(u)$ for capturing the relations between aspect entity $u$. 

\noindent \textbf{Cluster graph embedding}:
The weighted cluster graph $G_C(V_C, E_C, W_C)$ is a compact representation of the KG where each vertex $v \in V_C$ is a cluster of vertices (entities) of the knowledge graph $G=(V,E)$. We use the Louvain hierarchical graph clustering \citep{Blondel_2008} algorithm for clustering the entire knowledge graph.
Edge set $E_C$ is calculated as: $(i,j)\in E_C$, $\forall i,j \in V_C$ if there is a connected pair of KG entities from clusters $i$ and $j$. The weight between clusters $i$ and $j$, $W_C(i,j)$, is calculated as the fraction of actual edges between clusters $i$ and $j$ and the maximum edges possible between the two clusters, i.e, $|i|*|j|$, where $|i|$ is the number of nodes present in cluster $i$. 
We use a modified GraphSAGE embedding technique to construct the cluster embeddings $\bz_C(i)$, $i\in V_C$ of a weighted graph by optimizing the following graph based loss function: 
\begin{align*}
\resizebox{0.5\textwidth}{!}{$J_C(\boldsymbol{z}_C(i))  = -log(\sigma(\boldsymbol{z}_C(i)^T\boldsymbol{z}_C(j)))-
 Q \cdot \Bbb{E}_{k \sim P_n(j)}log(\sigma(-\boldsymbol{z}_C(i)^T \boldsymbol{z}_C(k)))$}
\end{align*}
where $\boldsymbol{z}_C(i)$ is the output representation of $i \in V_C$, $\sigma$ is the sigmoid function, $j \in V_C$ is a cluster co-occurring with $i$ on a fixed weighted random walk defined by $W_C(i,j)$, $P_n$ is the negative sampling distribution, $Q$ is the number of negative samples, $k \in V_C$ is a negative sample.

\noindent \textbf{Subgraph embedding}:
The vertex set $V_s$ of the entity-relation subgraph $G_s(V_s, E_s)$ consists of all aspect entities extracted from instances in the training dataset, while the edge set $E_s$ is the subset of induced edges from the original KG.
We use the standard GraphSAGE embedding and loss function to construct the \textit{subgraph similarity embedding}, $\bz_s(u)$ for aspect entity $u$.
To preserve the local neighbourhood information as well as global graph structure in the knowledge graph, we use the concatenation of subgraph and cluster graph embedings as our two level entity embedding: 
$\boldsymbol{z}(u)=[\boldsymbol{z}_C(i);\boldsymbol{z}_s(u)]$, where $u \in V_s$ and $i \in V_C$ such that $u$ is an entitiy in cluster $i$.
Figure \ref{Figure:2} shows the methods for aspect disambiguation and two-level entity embedding on the left side in the overall scheme. 

\subsection{ALSC with entity relation learning}
\label{sec:end-to-end}

In this section, we incorporate the concatenated entity embedding proposed above into two state-of-the-art ALSC models: TNet and BERT (described in section \ref{sec:background_alsc}). 
We propose two ways of incorporating the information contained in entity relations from KG into ALSC: (1) using static embeddings, and (2) by performing end-to-end learning.
For incorporation of static embedding in TNet, the final entity embedding $\boldsymbol{z}(u)$ for entity $u$ is concatenated with final layer CPT block output $\boldsymbol{h^{(L)}}$ as $\boldsymbol{h_{concat}^{(L)}}=[\boldsymbol{h^{(L)}};\boldsymbol{z}(u)]$ and this new aspect specific contextual representation $\boldsymbol{h_{concat}^{(L)}}$ is sent as input to the convolution layer module as described in section \ref{sec:background_alsc}. The final layers and loss function is same as TNet. We call this model \textbf{Aspect Relation-TNet} (\textbf{AR-TNet}).
We incorporate the entity embedding  $\boldsymbol{z}(u)$ into BERT by concatenating it with representation of CLS token $h_{[CLS]}$, as: $h_{[CLS]_{concat}} = [h_{[CLS]};\boldsymbol{z}(u)]$. 
Here $h_{[CLS]}$ is the final aspect-specific sentence representation for an ALSC instance taken from domain specific BERT model (BERT-ADA) \citep{DBLP:conf/lrec/RietzlerSOE20}, and further fine-tuned on ALSC task. 
We call this model \textbf{Aspect Relation-BERT} (\textbf{AR-BERT}). We also incorporate the static entity embedding $\boldsymbol{z}(u)$ into SDGCN-BERT \citep{DBLP:journals/kbs/ZhaoHW20}, in an analogous way to train the \textbf{Aspect Relation-BERT-S} (\textbf{AR-BERT-S}) model, through finetuning on ALSC. These models are referred to in Table \ref{Table:3} with \textbf{‘wo end-to-end’} in parenthesis because these models are trained without an end-to-end strategy. The end-to-end training of our proposed models \textbf{AR-BERT}, \textbf{AR-BERT-S}, and \textbf{AR-TNet} (for the base models BERT-ADA, SDGCN-BERT, and TNet, respectively) are discussed in the following section. 

\newcommand{\cL}{\mathcal{L}}

\noindent
\textbf{End-to-end learning}:
Incorporation of GraphSAGE embeddings into ALSC models provide minor improvements to polarity prediction, since the aspect embeddings are not fine-tuned for the ALSC task. This is achieved with end-to-end training of the aspect embedding and ALSC models. The architecture of our end-to-end models are same as the models proposed with static embeddings above. Hence, for BERT based models, we calculate the final embeddings for a sentence and aspect pair as: $h_{[CLS]_{concat}} = [h_{[CLS]};\boldsymbol{z}(u)]$, where $\boldsymbol{z}(u)=[\boldsymbol{z}_C(i);\boldsymbol{z}_s(u)]$. For TNet-based models, $\boldsymbol{h_{concat}^{(L)}}=[\boldsymbol{h^{(L)}};\boldsymbol{z}(u)]$. For both models, let $\cL_{ALSC}$ denote the loss incurred from  ALSC training, and $\cL_{GS}$ be the loss incurred from GraphSage using the subgraph $G_s=(V_s,E_s)$. We optimize the following loss for joint training:
\begin{align*}
    \cL_{joint}(\Theta_{ALSC}, \{ \bz_s(u) \}) = \alpha_1 \cL_{ALSC} + \alpha_2 \cL_{GS}
\end{align*}
where, $\Theta_{ALSC}$ are all the parameters from ALSC model, and $\{ \bz_s(u) \}$ are subgraph embeddings from $G_s$. We minimize the above loss w.r.t. $\Theta_{ALSC}, \{ \bz_s(u) \}$, while keeping $\bz_C(i)$ fixed to pre-learned GraphSAGE embeddings. 

\subsection{Incorrect Disambiguation Detection}
\label{sec:disambiguation-correction}

Many of the misclassifications using models like BERT-GS, are due to incorrect disambiguation of aspect entities (see section \ref{sec:Improvement using Semantic Probing}). 
In this section, we develop a scalable algorithm for identifying incorrect aspect disambiguations and mitigating their effect by setting the corresponding (modified) embedding to zero vector. We rely on the BERT aspect embedding vectors $h_{[CLS]}$ (called $h$ in this section for brevity) for the same. 
However, BERT embeddings encode many modalities of information including syntactic dependencies \citep{DBLP:conf/naacl/HewittM19}, semantic similarities, and entity relations \citep{DBLP:conf/nips/ReifYWVCPK19}.
We propose to use a learned similarity function  $\mathcal{S}_{B}(h_i,h_j)$ which captures the entity similarity between two BERT embeddings $h_i$ and $h_j$ of two entity mentions. Hence, following \citep{DBLP:conf/nips/ReifYWVCPK19}, we propose to use the following form of similarity function:
$$\mathcal{S}_{B}(h_i, h_j) = \sigma ((B \cdot h_i)^T(B \cdot h_j))$$
where, $B \in R^{dim_B*dim_h}$ is a learned parameter. The parameter $B$ can be thought of as a ``probing function'' \citep{DBLP:conf/nips/ReifYWVCPK19},  projecting BERT embedding $h$ into a space which only distills out the entity relations.


Algorithm \ref{alg:incorrect}, describes the steps for learning the probing function parameter $B$, which extracts entity relational similarities from BERT embeddings, and calculation of the modified embeddings.
The key idea is: \textit{aspects which are close in graph embedding space should also have high similarity of BERT embeddings}. The algorithm proceeds by constructing triplets $(i,j,k)$ of aspects where aspects $i$ and $j$ are closer, but $i$ and $k$ are not closer. It then learns $B$ by minimizing the loss: $\sum_{(i,j,k)\in \tau} ({\mathcal S_{B}(h_i, h_k)} - {\mathcal S_{B}(h_i, h_j)})$.  
For each aspect $i$, and for all it's top $n$ close aspects $j$ and rest far away aspects $k$, we modify it's corresponding concatenated entity embedding as follows: 
\begin{equation}
\resizebox{0.45\textwidth}{!}{$\boldsymbol{z}_{mod}(i)=\begin{cases}\{\vec{\scriptstyle 0}\}^{dim_h},& \text{if } {\mathcal S_{B}(h_i, h_j)} - {\mathcal S_{B}(h_i, h_k)} \geq 0\\ \boldsymbol{z}(i), & \text{otherwise}  \end{cases}$}
\label{eq:zmod}
\end{equation}
$\{\vec{\scriptstyle 0}\}^{dim_h}$ is the zero vector of dimension $dim_h$. We call ALSC models jointly (end-to-end) trained with these corrected embeddings as: \textbf{AR-BERT-idd}, \textbf{AR-BERT-S-idd}, and \textbf{AR-TNet-idd}; corresponding to base models BERT-ADA, SDGCN-BERT, and TNet. 

\newcommand{\cA}{\mathcal{A}}
\newcommand{\cD}{\mathcal{D}}

\begin{algorithm}
\caption{Incorrect Disambiguation Detection}
\label{alg:incorrect}
\begin{algorithmic}[1] \hrule
\REQUIRE Set of aspects $\cA(\cD)$ in Dataset $\cD$, Graph aspect embeddings $\{ \bz(u) \}, u\in \cA(\cD)$, 
 BERT aspect embeddings $\{ \vh(u) \}, u\in \cA(\cD)$ \\ 
 \ENSURE Probing distance fn. parameter: $B$,
 Corrected embedding: $\bz_{mod}(u)$\\ \hrule
 \STATE randomly initialize $B$
 \STATE LIST $\tau \leftarrow \phi$
 \FORALL{$i\in \cA(\cD)$}
 \STATE $\cA_i(\cD) \leftarrow$ sort other aspects  $\bz(u), u\in \cA(\cD)$ in decreasing order of distance to $\bz(i)$
 \STATE Sample $j$ from top $n$ in $\cA_i(\cD)$, and $k$ from rest; emit $\tau\leftarrow\tau \cup (i,j,k)$
 \ENDFOR
 \STATE learn $B$: $\min_B \sum_{(i,j,k)\in \tau} ({\mathcal S_{B}(h_i, h_k)} - {\mathcal S_{B}(h_i, h_j)}) + \lambda \norm{B}^2$
 \STATE Calculate $\bz_{mod}(u)$ using eqn. \ref{eq:zmod}
\end{algorithmic}
\end{algorithm}

\subsection{Multi-modal Explanation Generation}
\label{sec:multimodal-explanation} 

The ALSC models proposed above incorporate semantic and syntactic information from the text data, and aspect relations from the knowledge graphs. Since, the information from these two modes are combined using a deep neural network, for a given test example, the information content in each of the modes is not obvious. In this section, we describe a global posthoc explanation model for generating \textit{multi-modal explanations} for predictions provided by the proposed model architecture. For simplicity, we summarize the proposed architecture into 3 components: (1) the text feature extractor model for input context and aspect $\vec{x}$, $\boldsymbol{M_t}(\vec{x}): \boldsymbol{X} \rightarrow {h(\vec{x})}$, (2) the graph embedding model $\boldsymbol{M_g}(G): \boldsymbol{G} \rightarrow z(G)$, and (3) the final prediction model  $\boldsymbol{M_o}(h(\vec{x}), z(G)): [h(\vec{x});z(G)] \rightarrow \{P,N,O\}$. Hence, $h(\vec{x}) = h^{(L)}(\vec{x})$ for the TNET model and $h((\vec{x}) = h_{[CLS]}(\vec{x})$ for the BERT model. $z(G) = \mathbf{z(u)}$ (section \ref{sec:Target-Relation-Incorporation}) and $\boldsymbol{M_o}(h(\vec{x}), z(G))$ is a feedforward neural network with softmax output. 

For generating global explanations from text features $h((\vec{x})$ we build on the method proposed in \citep{yeh2019completeness}. Let $\theta_{con}$ be the matrix of $k$-concept vectors used for explaining salient features of prediction, and $\theta_{dec}$ be the parameters of the decoder network $g$, which reconstructs the original vectors from concept embeddings $\theta_{con}^T h(\vec{x})$. The parameters of the text explanation model $E_t$ ($\theta_{con},\theta_{dec}$) are learned by minimizing the regularized negative log-likelihood function:
\begin{align*}
\resizebox{0.45\textwidth}{!}{
$\cL_{{E_t}}(\theta_{con}, \theta_{dec}) = 
\mathbb{E}_{\vec{x}}[ - \log P(\boldsymbol{M_o}(g( \theta_{con}^T h(\vec{x}), \theta_{dec}),\ z(G))] + \mathbf{R}(\theta_{con})$}
\end{align*}
where, $R(\theta_{con})$ is the diversity regularity between concepts described in \cite{yeh2019completeness}. 
Explanation words are extracted by selecting top scoring words $x_i$ according to the score $\max_k (\theta_{con}^T h(\vec{x_i})$.
For the graph explanation model $E_g$ with parameters $\theta_g$, we use parameterized graph explainer (PGExplainer) model \citep{luo2020parameterized}, minimizing the entropy loss:
\begin{align*}
\resizebox{0.45\textwidth}{!}{
$\cL_{{E_g}}(\theta_g) = 
\mathbb{E}_{G_S \sim q(\theta_{g})} [H( \boldsymbol{M_o}(h(\vec{x}), \boldsymbol{M_g}(G) )\ |\ \boldsymbol{M_o}(h(\vec{x}), \boldsymbol{M_g}(G_S) ) )]$}
\end{align*}
where $G_S$ is explanation subgraph sampled from a distribution $q$ parameterized by $\theta_{g}$.

A key challenge in generating multi-modal explanations is to identify the significance of individual modes. To this end, we define the significance variables for graph and text mode $s_g,s_t\in[0,1]$, respectively. Given the explanations provided by unimodal explanation models, we generate the perturbed input text $\tilde{x}$ by removing the explanation words from $\vec{x}$, and the perturbed input graph $\tilde{G}$ by removing vertices and induced edges from the explanation subgraph. The significance labels are set as: $s_t=1$ if $\boldsymbol{M_o}(h(\tilde{x}), z(G)) \neq \boldsymbol{M_o}(h(\vec{x}), z(G))$ and $0$ otherwise, and analogously for $s_g$. We also train  feed-forward neural networks with sigmoid activation for predicting $s_t,s_g$ from inputs $\vec{x}), z(G)$: $s_t = S_t(\vec{x}), z(G))$, and $s_g= S_g(\vec{x}), z(G))$. Hence the joint multi-modal explanation loss is given as:
\begin{align*}
\resizebox{0.45\textwidth}{!}{
$\cL_{{mm}}(\theta_{con}, \theta_{dec} \theta_{g}, S_t, S_g) = S_t * \cL_{{E_t}} + S_g * \cL_{{E_g}} + \lambda( L(s_t,S_t) + L((s_g,S_g) )$}
\end{align*}
The first two terms are significance weighted explanation losses for individual modes, and the last two terms are binary cross-entropy losses for matching significance predictors to respective significance values generated by the unimodal explanation predictors. $\lambda$ controls the weightage given to initial unimodal significance values.


\section{Experiments}

In this section, we report experimental results to empirically ascertain whether the proposed models indeed perform better than the existing state of the art methods.

\subsection{Experimental Setup}


\textbf{Datasets and baselines}: We evaluate our proposed models on the three benchmark datasets: LAPTOP and REST datasets from SemEval 2014 Task 4 subtask 2 \citep{DBLP:conf/semeval/PontikiGPPAM14} which contains reviews from Laptop and Restaurant domain respectively and the TWITTER dataset \citep{DBLP:conf/acl/DongWTTZX14} containing Twitter posts. 
For TNet-based models, we perform the same prepossessing procedure as done in \citep{DBLP:conf/acl/LamLSB18}.
 We compare results of our proposed models with state-of-the-art methods reported in table \ref{Table:3}.

\noindent \textbf{Aspect disambiguation and KG Embedding}: 
For each aspect in the dataset $\mathcal{D}$ mentioned above, we disambiguate its corresponding entity in the knowledge graph using the \textit{wikifier} API \citep{DBLP:journals/informaticaSI/BrankLG18}. 
We use hierarchical Louvain graph clustering \citep{Blondel_2008} algorithm for clustering the KG and constructing the weighted cluster graph $G_C(V_C, E_C, W_C)$(ref section \ref{sec:two-level entity embedding}). 
Statistics of the knowledge graph and its corresponding cluster graph and sub-graphs are shown in Table \ref{Table:2}.
For training entity sub-graph and weighted cluster graph embedding, we use GraphSAGE mean as aggregate function.
For training GraphSage \citep{DBLP:conf/nips/HamiltonYL17}, we sample 25 nodes for layer 1 and 10 nodes for layer 2 using a random walk. The output hidden representation dimension is set as 50, and the number of negative samples $Q$ taken as 5. Default values are used for all other parameters.

\noindent \textbf{ALSC and probing function training}: For TNet-based models, we used 20\% randomly held-out training data as the development set. We train the model for 100 epochs and select the model corresponding to the maximum development-set accuracy. Following \citep{DBLP:conf/acl/LamLSB18}, we use the same set of hyperparameters and report the maximum accuracy obtained on the test set over multiple runs.
For training of BERT-based models, we use the procedure suggested in \citet{DBLP:conf/lrec/RietzlerSOE20}, for both pre-training and fine-tuning. 
For end-to-end training of ALSC with entity embedding generation, we use Adam optimizer with a learning rate of $3\cdot10^{-5}$, batch size of 512 for GraphSAGE-based entity embedding generation and 32 for ALSC task, number of epochs as 7. All the other hyper-parameters in GraphSAGE based entity embedding generation and ALSC task follow the same values in individual training. 
For training of the probing function $B$, we use Adam optimizer with a learning rate of $1\cdot10^{-5}$, the batch size of 128, the number of epochs as 100, the probe dimension $dim_B$ as 100, and the regularization rate as 0.01.
We use the same evaluation procedure suggested in \citet{DBLP:conf/lrec/RietzlerSOE20}, e.g. we conducted 9 runs with different initializations for all the experiments, and reported the average performance on the given test set. The model with the best training error over 7 epochs is taken as the final model for all runs.

\begin{table}[h]
    \caption{Statistics of knowledge graph, weighted cluster graph and entity relation sub-graphs.}
    \label{Table:2}
    \centering 
    {\small
    \begin{tabular}{llll} \hline
         \multicolumn{4}{l}{\textbf{Knowledge Graph Embedding} } \\ \hline
           \#Edges & \#Nodes & \#Clusters &
           Max. inter \\
           & & & -cluster degree \\ \hline
         173068197 & 22504204 & 606 & 341 \\ \hline \hline  
         \multicolumn{4}{l}{\textbf{Sub-graph Embedding}}\\ \hline
         Dataset & \#Nodes & \#Edges & Max. Node \\ 
         & & & Degree \\ \hline
           LAPTOP & 785 & 4477 & 107 \\ 
           REST & 1031 & 7305 & 136 \\ 
           TWITTER & 120 & 429 & 40 \\ \hline 

    \end{tabular}%
    }
\end{table}

\begin{table*}[t]
\caption{Experiment results on various datasets(\%). The marker * refers to p-value \textless 0.01 when comparing
with respective baselines. \% in bracket of best performing models implies overall gain wrt. its' baselines.}
\label{Table:3}
\centering
{\small
\begin{tabular}{lllllll} \hline
\textbf{Model} & \multicolumn{2}{c}{LAPTOP} & \multicolumn{2}{c}{REST} & \multicolumn{2}{c}{TWITTER}\\
& ACC &  Macro-F1 & ACC &  Macro-F1 & ACC &  Macro-F1\\ \hline \hline 
\multicolumn{7}{l}{\textbf{Baseline models for ALSC}} \\ \hline
TNet \citep{DBLP:conf/acl/LamLSB18} & 76.33 & 71.27 & 79.64 & 70.20 & 78.17 & 77.17\\
BERT-base \citep{DBLP:conf/naacl/DevlinCLT19} & 77.69 & 72.60 & 84.92 & 76.93 & 78.81 & 77.94 \\
SDGCN-BERT \citep{DBLP:journals/kbs/ZhaoHW20} & 81.35 & 78.34 & 83.57 & 76.47 & 78.54 & 77.72 \\
BERT-ADA \citep{DBLP:conf/lrec/RietzlerSOE20} & 80.25 & 75.77 & 87.89 & 81.05 & 78.90 & 77.97 \\ \hline \hline
\multicolumn{7}{l}{\textbf{ALSC with Aspect relation incorporation}} \\ \hline
AR-TNet & 78.80$^\star$ & 73.87$^\star$ & 83.40$^\star$ & 73.91$^\star$ & 80.52$^\star$ & 79.79$^\star$ \\
AR-BERT & 81.73$^\star$ & 77.07$^\star$ & 89.38$^\star$ & 82.47$^\star$ & 80.91$^\star$ & 80.15$^\star$ \\ 
AR-BERT-S & 82.37$^\star$ & 79.21$^\star$ & 85.27$^\star$ & 78.07$^\star$ & 79.67$^\star$ & 78.89$^\star$ \\ 
\hline \hline
\multicolumn{7}{l}{\textbf{ALSC with Aspect relation and incorrect disambiguation detection}} \\ \hline
AR-TNet-idd & 80.09$^\star$ & 75.11$^\star$ & 84.64$^\star$ & 75.17$^\star$ & 81.64$^\star$ & 80.84$^\star$ \\ \hline
AR-BERT-idd & 82.91$^\star$ & 78.31$^\star$ & \textbf{90.62}$^\star$ &  \textbf{83.81}$^\star$ & \textbf{82.08}$^\star$ & \textbf{81.21}$^\star$ \\
& & &  (+3.11\%) & (+3.40\%) & (+4.03\%) & (+4.15\%)\\
\hline
AR-BERT-S-idd & \textbf{83.62$^\star$} & \textbf{80.43$^\star$} & 86.61$^\star$ & 79.37$^\star$ & 80.86$^\star$ & 80.03$^\star$ \\
& (+2.79\%) & (+2.67\%) & & & & \\
\hline \hline
\multicolumn{7}{l}{\textbf{Results with explanation concepts}} \\ \hline
AR-BERT-idd ($h(x)=g( \theta_{con}^T h(\vec{x}), \theta_{dec})$) & - & - & 90.18 & 83.39 & 81.50 & 80.64 \\
AR-BERT-S-idd ($h(x)=g( \theta_{con}^T h(\vec{x}), \theta_{dec})$)& 83.38 & 80.25 & - & - & - & - \\ 
AR-BERT-idd ($G=G_{S}$)  & - & - & 90.36 & 83.57 & 81.94 & 81.07 \\
AR-BERT-S-idd ($G=G_{S}$ ) & 83.54  & 80.41  & - & - & - & - \\ 
\hline \hline
\multicolumn{7}{l}{\textbf{Results with ablations of explanations}} \\ \hline
AR-BERT-idd (with $\Tilde{x}$) & - & - & 34.13 &  33.25 & 33.81 & 33.17 \\
AR-BERT-S-idd (with $\Tilde{x}$) & 34.48 & 32.43 & - & - & - & - \\
AR-BERT-idd (with $\Tilde{G}$) & - & - & 88.01 &  81.11 & 78.75 & 77.67 \\
AR-BERT-S-idd (with $\Tilde{G}$) & 81.09 & 78.15 & - & - & - & - \\
\hline \hline
\end{tabular}
}
\end{table*}

\begin{table}[h]
    \caption{Confusion matrices of predictions of AR-TNet-idd vs TNet and AR-BERT-idd vs BERT-ADA w.r.t. correct and incorrect classification}
    \label{Table:4}
    \centering
    {\small
    \begin{tabular}{c|cc|cc} \hline
          \textbf{Baseline} & \multicolumn{2}{c}{\textbf{AR-TNet}} & \multicolumn{2}{c}{\textbf{AR-BERT}} \\ 
          & \multicolumn{2}{c}{\textbf{[idd]}} & \multicolumn{2}{c}{\textbf{[idd]}} \\ \hline \hline
          \textbf{Prediction}  & Correct & Incorrect & Correct & Incorrect \\ \hline
         \multicolumn{5}{l}{\textbf{LAPTOP}} \\ \hline \hline
         Correct & 487 & 0 & 512 & 0 \\ 
         Incorrect & 24 & 127 & 17 & 109\\ \hline
         \multicolumn{5}{l}{\textbf{REST}} \\ \hline \hline
         Correct & 892 & 0 & 984 & 0 \\
         Incorrect & 56 & 172 & 31 & 105 \\ \hline
         \multicolumn{5}{l}{\textbf{TWITTER}} \\ \hline \hline
         Correct & 541 & 0 & 547 & 0 \\
         Incorrect & 24 & 127 & 22 & 124 \\ \hline
    \end{tabular}%
    }
\end{table}

\begin{table}[h]
    \caption{Fraction of incorrectly predicted examples in disambiguation categories.}
    \label{Table:5}
    \centering
    {\small
    \begin{tabular}{cccc} \hline
          & \textbf{\# i\_c / \# u\_d} & \textbf{\# i\_c / \# i\_d} & \textbf{\# i\_c / \# c\_d} \\ \hline
          \multicolumn{4}{l}{\textbf{AR-BERT}}\\ \hline \hline
          LAPTOP & 8 / 10 & 47 / 53 & 62 / 575  \\
          REST & 12 / 14 & 76 / 87 & 31 / 1019 \\
          TWITTER & 2 / 2 & 8 / 14 & 122 / 676 \\ \hline
          \multicolumn{4}{l}{\textbf{AR-BERT-idd}}\\ \hline \hline
          LAPTOP & 6 / 10 & 41 / 53 & 62 / 575  \\
          REST & 10 / 14 & 65 / 87 & 30 / 1019 \\
          TWITTER & 0 / 2 & 0 / 14 & 124 / 676 \\ \hline
    \end{tabular}%
    }
\end{table}

\begin{table*}[t]
    \caption{Examples of mistakes by BERT-ADA which were correctly predicted by AR-BERT-idd. Red and green backgrounds indicate context explanations from multi-modal explanation model for BERT-ADA and AR-BERT-idd.}
    \label{Table:6}
    \centering
    {\small
    \begin{tabular}{llll} \hline
         \textbf{Sentence} & \textbf{BERT} & \textbf{AR-BERT} & \textbf{Graph-mode} \\ 
         & \textbf{-ADA} & \textbf{-idd} & \textbf{Explanation} \\ \hline \hline
         \multicolumn{4}{l}{\textbf{LAPTOP}} \\ \hline \hline
         However, I can \mybox[fill=green!20]{refute} that \textbf{[OSX]\textsubscript{NEG}} is \mybox[fill=red!20]{"FAST"}. & \textbf{POS} & \textbf{NEG} & \textbf{Hard\_disk\_drive (39)} \\ \hline
         From the speed to the multi touch gestures this operating system \mybox[fill=green!20]{beats} \textbf{[Windows]\textsubscript{NEG}} \mybox[fill=red!20]{easily}. & \textbf{POS} & \textbf{NEG} & \textbf{Software (64)}\\ \hline
         \mybox[fill=green!20]{Did not enjoy} the new \textbf{[Windows 8]\textsubscript{NEG}} and touchscreen functions. & \textbf{NEG} & \textbf{NEG} & - \\ \hline \hline
         \multicolumn{4}{l}{\textbf{REST}} \\ \hline \hline
         \mybox[fill=green!20]{Anywhere else}, the \textbf{[prices]\textsubscript{POS}} would be 3x as \mybox[fill=red!20]{high}! &  \textbf{NEG} & \textbf{POS} & \textbf{Service (140)} \\ \hline
         A \mybox[fill=red!20]{beautiful} \mybox[fill=green!20]{atmosphere}, perfect for \textbf{[drinks]\textsubscript{NEU}} and/or appetizers.  &  \textbf{POS} & \textbf{NEU} & \textbf{Food (365)} \\ \hline 
         The \textbf{[bread]\textsubscript{POS}} is \mybox[fill=green!20]{top notch} as \mybox[fill=green!20]{well}. & \textbf{POS} & \textbf{POS} & - \\ \hline \hline
         \multicolumn{4}{l}{\textbf{TWITTER}} \\ \hline \hline
         noobus \mybox[fill=green!20]{Turns} out \textbf{[Snoop Dogg]\textsubscript{NEG}} is actually \mybox[fill=red!20]{pretty} \mybox[fill=green!20]{funny} . & \textbf{POS} & \textbf{NEG} & \textbf{Barack\_Obama (343)} \\ \hline 
         \mybox[fill=green!20]{just} got \mybox[fill=green!20]{hold} of an \textbf{[Ipod]\textsubscript{NEU}} . . it will be \mybox[fill=red!20]{fun} learning how to use it on the bus trip to & \textbf{POS} & \textbf{NEU} & \textbf{IPhone (168)} \\ canberra this monday & & \\ \hline
         Is it just me , or does \textbf{[John Boehner]\textsubscript{NEG}} \mybox[fill=red!20]{sound} like a \mybox[fill=red!20]{newsman} ? & \textbf{NEU} & \textbf{NEU} & - \\
         Sounds like he belongs on CBS Nightly \mybox[fill=red!20]{News} . & & &  \\ \hline
         
    \end{tabular}
    }
\end{table*}

 \subsection{Comparison of ALSC Models}
\label{sec:comparison}
 
 Table \ref{Table:3} reports all baseline and proposed models' performance, using two standard metrics: macro-accuracy (\textbf{ACC}) and macro-F1 score (\textbf{Macro-F1}). 
We observe that AR-BERT-idd outperforms all the other models on the REST and TWITTER dataset, and AR-BERT-S-idd outperforms all the other models for the LAPTOP dataset, in terms of both Accuracy and Macro-Averaged F1 scores. 
Hence, we can conclude that \textit{representation of relations between aspect entities helps in training better ALSC models}.
The improvement in the performance by the proposed AR-BERT-idd and AR-BERT-S-idd models over other BERT-based baseline models imply that the DBpedia knowledge graph encodes information which supplements the information contained in BERT embeddings of the aspect terms.
 We also note that BERT based baseline models, e.g. BERT-ADA, perform better than other models, e.g. TNET, as they utilize the context-sensitive word embeddings fine-tuned on domain-related datasets. 
 
\begin{figure}[h]
    \centering
    {\small
    \includegraphics[width=6cm]{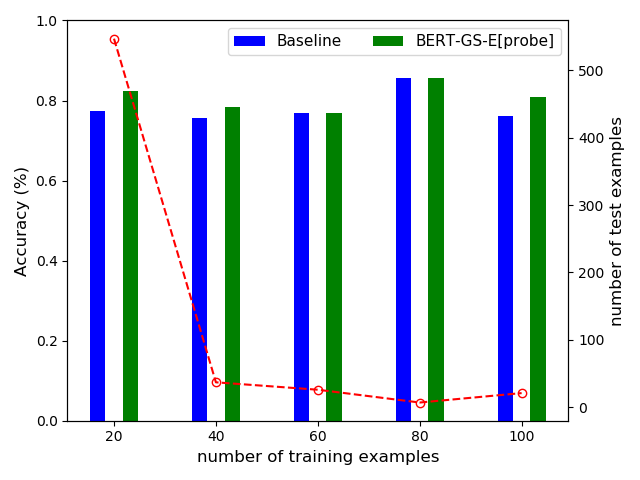}
    \caption{Effect of number of training datapoints}
    \label{fig:fig3}
    }
\end{figure}

\noindent \textbf{Effect of training data scarcity}: Figure \ref{fig:fig3} reports the accuracy of the baseline model (BERT-ADA, blue bar) and the proposed model (green bar),  for all test aspects in the LAPTOP dataset. The test aspects are bucketed according to their training data counts, and the bars report average accuracy for all aspects in the buckets.
We can see that for aspects that have 0 -- 20 training points, the proposed method outperforms the baseline. Hence, we conclude that for aspects with a low number of training data points, the proposed method improves the performance of ALSC by borrowing information from nearby aspects in the KG. The red line shows the number of test data points for each of the buckets. We find that a large fraction of test aspects have fewer than 20 training data points.


\noindent 
\textbf{Error Analysis}: Table \ref{Table:4} shows the confusion matrices of predictions of AR-TNet-idd and AR-BERT[idd] w.r.t. their respective baseline models on the three datasets. The top-left and bottom-right values report the number of correctly classified or misclassified examples by both methods in each sub-matrix. We can see that the proposed models do not induce any new errors which were not present in the respective baselines. Finally, we see that the bottom left entries in each table that report the new corrects (examples classified wrongly by existing methods but are classified correctly by the proposed methods) are much higher. Thus, we conclude that the new technique is an improvement over the old methods.

\noindent \textbf{Anecdotal examples}:
Table \ref{Table:6} illustrates a few examples misclassified by BERT-ADA and correctly predicted by AR-BERT-idd. Aspects in the sentences are marked in the bracket with corresponding sentiment labels in subscript. Context words captured by the models for corresponding predictions are extracted using the context explainer of multi-modal explanation generation model, and the most important aspect entity (if any) is extracted using graph explainer of multi-modal explanation model (for instances where both context and graph information is required to predict sentiment polarity correctly, the most important aspect entity is the node in the explanation sub-graph with highest training example). Context explanations verify that BERT-ADA captures context words that are always semantically related to the aspects e.g. `high' w.r.t. aspect `price', whereas AR-BERT-idd considers aspect as an entity and tries to captures context words associated with that entity or the most important aspect entity from the given context. For each dataset, the distribution of test examples in each mode of importance of multi-modal explanation is given in Table \ref{Table:8}.       

\subsection{Incorrect Disambiguation Detection} 
\label{sec:Improvement using Semantic Probing} 
In this section, we demonstrate the effectiveness of our probing function for incorrect disambiguation detection.
We categorized the aspects into 3 categories based on the disambiguation by wikifier: (1) \textbf{unknown (unk)} where there was no entity found, (2) \textbf{correct disambiguation (cd)} where the disambiguated aspect was mapped to the correct entity, and (3) \textbf{incorrect disambiguation (id)} where the disambiguated aspect was mapped to an incorrect entity, based on manual annotation.
Table \ref{Table:5} shows the number of incorrectly classified \textbf{(ic)} examples (by the ALSC model) in each disambiguation category out of the total number of examples in that category ($\# ic / \# Total)$.
We see that compared to AR-BERT (wo end-to-end) and AR-BERT, AR-BERT-idd has significantly fewer incorrectly classified for the \textbf{unknown} and \textbf{incorrect disambiguation} categories. For the \textbf{correct disambiguation} category, all methods have the similar fraction of misclassification, which is much lower than the other two categories.

\begin{figure}[h]
    \centering
    {\small
    \includegraphics[width=6cm]{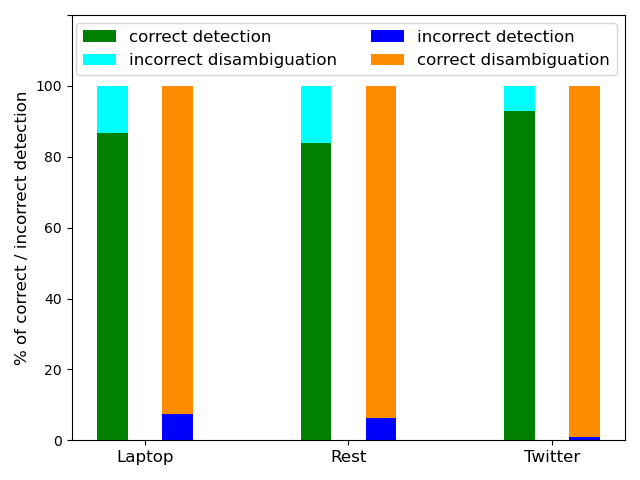}
    \caption{Percentage of correct / incorrect detection of disambiguation}
    \label{fig:fig2}
    }
\end{figure}

Figure \ref{fig:fig2} reports the percentage \textbf{i\_d} which were detected correctly (left bar), and the \textbf{c\_d} which were marked incorrectly (right bar) by the probing scheme. It can be seen that more than 80\% of \textbf{i\_d} examples have been correctly detected, and less than 5\% of \textbf{c\_d} examples have been wrongly flagged. Hence, we conclude that our incorrect disambiguation detection method shows excellent performance, while also being highly scalable.

\begin{table}[h]
    \caption{Distribution of test examples in each mode of multi-modal explanation with mode importance}
    \label{Table:8}
    \centering
    {\small
    \begin{tabular}{l|lll} \hline 
          \textbf{Important} & LAPTOP & REST & TWITTER \\ 
          \textbf{Mode} & & & \\ \hline \hline
          \textbf{Text} ($S_t=1$ and $S_g=0$) & 300 (0) & 588 (0) & 355 (0) \\ \hline
          \textbf{Graph} ($S_t=0$ and $S_g=1$) & 34 (0) & 45 (0) & 49 (0) \\ \hline
          \textbf{Both} ($S_t=1$ and $S_g=1$) & 234 (15) & 427 (31) & 216 (22) \\ \hline 
          \textbf{None} ($S_t=0$ and $S_g=0$)& 70 (0) & 60 (0) & 72 (0) \\  \hline \hline
          \textbf{Total} & 638 & 1120 & 692 \\ \hline
    \end{tabular}%
    }
\end{table}

\subsection{Effectiveness of Multi-modal Explanations} 
\label{sec:explanation ablation study}

In this section, we report experimental results studying the effectiveness of multi-modal explanations predicted by the proposed model. Table \ref{Table:3}, section \emph{Results with explanation concepts} reports the predictive performance of ALSC models with explanations as input. The top two rows use the text concept embeddings as input, while the next two rows use the explanation subgraph as the input. We note that in all these cases the prediction performance remains similar to the original model, with the maximum accuracy difference of $<0.5\%$ and maximum macro-F1 score difference of $<1.5\%$. This demonstrates the effectiveness of the explanation models. We also notice that predicting using the text concept embeddings results in marginally lower performance compared to predicting using graph explanation, which is expected due to the overall higher importance of the text mode. Table \ref{Table:3}, section-\emph{Results with ablations with explanations} further demonstates the effectiveness of explanations by reporting the performance of the ALSC model with perturbations $\tilde{x}$ and $\tilde{G}$ removing the explanations of text and graph modes as input. The results (top two rows) clearly demonstrate that the text mode explanation include an important portion of the text, whose removal causes drastic drop in model performance.

Table \ref{Table:8} analyzes the effectiveness of prediction of important modes of explanation. Text mode implies $S_t=1$ and $S_g=0$, and other modes are defined analogously. We note that "Text" and "Both" mode cover  a majority of the test examples, which is expected. Graph mode covers a miniscule number of examples for which the ALSC prediction can be made by simply observing the aspect graph nodes. Finally, we observe a small number of examples for which the explanation model predicts that none of the modes are important. A majority of these examples are ones where the ALSC model erroneously predicts the ALSC labels, due to which there is no possibility of improvement using any of the modes.
The numbers in brackets in table \ref{Table:8} show the number of cases where there has been an improvement in label prediction by the ALSC model due to incorporation of aspect relations from graphs. Note that all the improvements have been in the cases of examples for which the explanation model predicts that both modes are important. This further demonstrates the effectiveness of the mode prediction models. 
Table \ref{Table:6} reports anecdotal examples from the three datasets along with text mode and graph mode explanations provided by the explanation model. Examples for each dataset shows examples with improvement in ALSC prediction due to incorporation of graphs. The entities corresponding to aspect graph nodes, and count of neighboring nodes in the aspect graph are shown in the graph mode explanation, which are intuitive.

\section{Conclusions}
In this paper, we present a scalable technique for incorporating aspect relations from large knowledge graphs, into state of the art deep learning based ALSC models. The resulting algorithm - AR-BERT, along with a novel incorrect disambiguation detection technique, results in consistent and significant improvements in ALSC performance on all benchmark datasets. This work also reports the first algorithm for multi-modal explanation generation across textual and graph data.

\bibliographystyle{ACM-Reference-Format}
\bibliography{sample-base}

\clearpage

\appendix

\section{APPENDIX}

\begin{table}[t]
\caption{Statistics of datasets.}
\label{Table:9}
\centering
{\small
\begin{tabular}{cccc}\hline
& \# Positive & \# Negative & \# Neutral  \\ \hline
\multicolumn{4}{l}{LAPTOP} \\ \hline \hline
Train & 987 & 866 & 460\\
Test & 341 & 128 & 169 \\ \hline
\multicolumn{4}{l}{REST} \\ \hline \hline
Train &2164 & 805 & 633\\
Test & 728 & 196 & 196 \\ \hline
\multicolumn{4}{l}{TWITTER} \\ \hline \hline
Train & 1567 & 1563 & 3127 \\
Test & 174 & 174 & 346 \\ \hline
\end{tabular}%
}
\end{table}

\begin{table}[t]
\caption{Performance comparison of our proposed models with other competitive models.}
\label{Table:10}
\centering
{\small
\begin{tabular}{lll}\hline
Model & LAPTOP & REST  \\ \hline
LSA+DeBERTa-V3-Large & \textbf{86.21} & \textbf{91.07} \\
LCF-ATEPC & 82.29 & 90.18 \\
ABSA-DeBERTa & 82.76 & 89.46 \\
RoBERTa+MLP & 83.78 & 87.37 \\
KaGRMN-DSG & 81.87 & 87.35 \\ \hline
AR-BERT-idd & 82.91 & 90.62 \\ 
AR-BERT-S-idd & 83.62 & 86.61 \\ \hline

\end{tabular}%
}
\end{table}

\begin{table*}[t]
\caption{Extended experiment results on various datasets(\%). The marker * refers to p-value \textless 0.01 when comparing
with respective baselines. \% in bracket of best performing models implies overall gain wrt. its' baselines.}
\label{Table:11}
\centering
{\small
\begin{tabular}{lllllll} \hline
\textbf{Model} & \multicolumn{2}{c}{LAPTOP} & \multicolumn{2}{c}{REST} & \multicolumn{2}{c}{TWITTER}\\
& ACC &  Macro-F1 & ACC &  Macro-F1 & ACC &  Macro-F1\\ \hline \hline 
\multicolumn{7}{l}{\textbf{Baseline models for ALSC}} \\ \hline
TNet \citep{DBLP:conf/acl/LamLSB18} & 76.33 & 71.27 & 79.64 & 70.20 & 78.17 & 77.17\\
TNet-ATT \citep{DBLP:conf/acl/TangLSGSSL19} & 77.62 & 73.84 & 81.53 & 72.90 & 78.61 & 77.72\\ 
BERT-base \citep{DBLP:conf/naacl/DevlinCLT19} & 77.69 & 72.60 & 84.92 & 76.93 & 78.81 & 77.94 \\
SDGCN-BERT \citep{DBLP:journals/kbs/ZhaoHW20} & 81.35 & 78.34 & 83.57 & 76.47 & 78.54 & 77.72 \\
BERT-ADA \citep{DBLP:conf/lrec/RietzlerSOE20} & 80.25 & 75.77 & 87.89 & 81.05 & 78.90 & 77.97 \\ \hline \hline
\multicolumn{7}{l}{\textbf{ALSC with Aspect relation incorporation}} \\ \hline
AR-TNet (wo end-to-end)  & 77.89$^\star$ & 72.96$^\star$ & 82.31$^\star$ & 72.97$^\star$ & 79.68$^\star$ & 78.83$^\star$ \\
AR-TNet & 78.80$^\star$ & 73.87$^\star$ & 83.40$^\star$ & 73.91$^\star$ & 80.52$^\star$ & 79.79$^\star$ \\
AR-BERT (wo end-to-end) & 80.87$^\star$ & 76.13$^\star$ & 88.21$^\star$ & 81.45$^\star$ & 79.83$^\star$ & 79.02$^\star$ \\
AR-BERT & 81.73$^\star$ & 77.07$^\star$ & 89.38$^\star$ & 82.47$^\star$ & 80.91$^\star$ & 80.15$^\star$ \\ 
AR-BERT-S (wo end-to-end) & 81.82$^\star$ & 78.75$^\star$ & 84.64$^\star$ &  77.34$^\star$ & 79.06$^\star$ & 78.36$^\star$ \\
AR-BERT-S & 82.37$^\star$ & 79.21$^\star$ & 85.27$^\star$ & 78.07$^\star$ & 79.67$^\star$ & 78.89$^\star$ \\ 
\hline \hline
\multicolumn{7}{l}{\textbf{ALSC with Aspect relation and incorrect disambiguation detection}} \\ \hline
AR-TNet-idd & 80.09$^\star$ & 75.11$^\star$ & 84.64$^\star$ & 75.17$^\star$ & 81.64$^\star$ & 80.84$^\star$ \\ \hline
AR-BERT-idd & 82.91$^\star$ & 78.31$^\star$ & \textbf{90.62}$^\star$ &  \textbf{83.81}$^\star$ & \textbf{82.08}$^\star$ & \textbf{81.21}$^\star$ \\
& & &  (+3.11\%) & (+3.40\%) & (+4.03\%) & (+4.15\%)\\
\hline
AR-BERT-S-idd & \textbf{83.62$^\star$} & \textbf{80.43$^\star$} & 86.61$^\star$ & 79.37$^\star$ & 80.86$^\star$ & 80.03$^\star$ \\
& (+2.79\%) & (+2.67\%) & & & & \\
\hline \hline
\multicolumn{7}{l}{\textbf{Results with explanation concepts}} \\ \hline
AR-BERT-idd ($h(x)=g( \theta_{con}^T h(\vec{x}), \theta_{dec})$) & - & - & 90.18 & 83.39 & 81.50 & 80.64 \\
AR-BERT-S-idd ($h(x)=g( \theta_{con}^T h(\vec{x}), \theta_{dec})$)& 83.38 & 80.25 & - & - & - & - \\ 
AR-BERT-idd ($G=G_{S}$)  & - & - & 90.36 & 83.57 & 81.94 & 81.07 \\
AR-BERT-S-idd ($G=G_{S}$ ) & 83.54  & 80.41  & - & - & - & - \\ 
\hline \hline
\multicolumn{7}{l}{\textbf{Results with ablations of explanations}} \\ \hline
AR-BERT-idd (with $\Tilde{x}$) & - & - & 34.13 &  33.25 & 33.81 & 33.17 \\
AR-BERT-S-idd (with $\Tilde{x}$) & 34.48 & 32.43 & - & - & - & - \\
AR-BERT-idd (with $\Tilde{G}$) & - & - & 88.01 &  81.11 & 78.75 & 77.67 \\
AR-BERT-S-idd (with $\Tilde{G}$) & 81.09 & 78.15 & - & - & - & - \\
\hline \hline
\end{tabular}
}
\end{table*}

\begin{table*}[t]
    \caption{Examples of Multi-modal Explanation Extraction. Aspect in the parenthesis, actual label in the subscript, predicted label around the aspect with proper color code: green for positive, red for negative and yellow for neutral, context explanation tokens are in blue color.}
    \label{Table:12}
    \centering
    {\small
    \begin{tabular}{lll} \hline
         \textbf{Sentence} & \textbf{Aspect} & \textbf{Explanation} \\ 
         & \textbf{-Entity} & \textbf{Node} \\ \hline \hline
         \multicolumn{3}{l}{\textbf{Mode Importance: Context}} \\ \hline \hline
         \mybox[fill=green!20]{\textbf{[Boot time]\textsubscript{POS}}} is \mybox[fill=blue!20]{super fast}, around anywhere from 35 seconds to 1 minute.
          & - & - \\ \hline
          The \mybox[fill=green!20]{\textbf{[bread]\textsubscript{POS}}} is \mybox[fill=blue!20]{top notch} as \mybox[fill=blue!20]{well}. & - & - \\ \hline
          i \mybox[fill=blue!20]{like} \mybox[fill=green!20]{\textbf{[Britney Spears ]\textsubscript{POS}}} new song ... i wan na hear it now = -LRB- & - & - \\ \hline \hline
         \multicolumn{3}{l}{\textbf{Mode Importance: Both}} \\ \hline \hline
         \mybox[fill=blue!20]{Did not enjoy} the new Windows 8 and \mybox[fill=red!20]{\textbf{[touchscreen functions]\textsubscript{NEG}}}. & \textbf{Touchscreen} & \textbf{Software(64)} \\ \hline
         Did I mention that the \mybox[fill=green!20]{\textbf{[coffee]\textsubscript{POS}}} is \mybox[fill=blue!20]{OUTSTANDING}? & \textbf{Coffee} & \textbf{Drink (35)} \\ \hline
         
         RT jaimemorelli : I would love to see a nuanced comparison of \mybox[fill=blue!20]{\textbf{[Google]\textsubscript{NEU}}} television & \textbf{Google} & \textbf{iPhone (168)} \\ 
         vs . hooking up my television to a Mac Mini and buying a wireless keyboard & & \\ \hline \hline
         \multicolumn{3}{l}{\textbf{Mode Importance: Graph}} \\ \hline \hline
         I would have given it 5 starts was it not for the fact that it had \mybox[fill=yellow!20]{\textbf{[Windows 8]\textsubscript{NEG}}} & \textbf{Windows\_8} & \textbf{Software(64)}\\ \hline 
         The \mybox[fill=red!20]{\textbf{[food]\textsubscript{NEU}}} did take a few extra minutes to come, but the cute waiters' jokes  & \textbf{Food} & \textbf{Food (365)} \\ 
         and friendliness made up for it. & & \\ \hline
         \mybox[fill=yellow!20]{\textbf{[Shaquille O'Neal]\textsubscript{NEG}}} to miss 3rd straight playoff game | & \textbf{Shaquille\_O'Neal} & \textbf{Los\_Angeles\_Lakers (157)} \\
         The ... : shaquille o'neal will miss his third straight play & & \\ \hline
         
    \end{tabular}
    }
\end{table*}

\subsection{Data Statistics}
The statistics of the three datasets are given in Table \ref{Table:9}.

\subsection{Implementation Details}
Our implementation is based on Tensorflow 1.X and all our experiments were executed on Quadro P5000 Single core GPU (16278MiB) with CUDA version: 10.0. 

\subsubsection{Aspect disambiguation}
We pass the sentence to the \href{http://www.wikifier.org/annotate-article}{Wikifier} using `POST' method and extract the entity from the annotation of the maximum sub-string of the aspect. If no annotation is found we link the retrieved entity as `Unknown'. 

\subsubsection{Knowledge Graph Clustering}
We extract the maximum weakly connected component of the DBPedia Knowledge Graph (KG) by converting the KG into a SNAP undirected graph\footnote{https://snap.stanford.edu/snappy/doc/reference/graphs.html} (fraction of nodes in maximum weakly connected component:
0.99983). We then cluster the extracted maximum weakly connected component using  hierarchical Louvain graph clustering algorithm. This algorithm returns a hierarchy tree with 5 levels in less than 40 minutes and the nodes in each level is as follows:
\begin{itemize}
    \item level 0: 22504204 nodes
    \item level 1: 227550 nodes
    \item level 2: 2938 nodes
    \item level 3: 769 nodes
    \item level 4: 606 nodes
\end{itemize}

\subsubsection{Domain Specific BERT pre-training}
Similar to \citep{DBLP:conf/lrec/RietzlerSOE20}, we pre-train the BERT model on the two domain specific publicly available datasets: Amazon electronics reviews\footnote{https://nijianmo.github.io/amazon/index.html} for LAPTOP domain and Yelp restaurants dataset\footnote{https://www.yelp.com/dataset/download} for REST domain. We adopt the similar pre-processing steps of \citep{DBLP:conf/lrec/RietzlerSOE20} by removing reviews with less than 2 sentences to enable Next Sentence Prediction (NSP) task of BERT language model. We also removes the reviews from Amazon review data which appear in the LAPTOP dataset to eliminate training bias towards those reviews. After pre-processing, we get around 1 million reviews for LAPTOP domain and we sample around 10 million reviews from pre-processed Yelp review dataset for REST domain. We run the pre-training of BERT model for 30 epochs and 3 epochs for LAPTOP and REST domain respectively to train the language model with significant large amount of data (equal for both domains).

\subsection{Other Competitive ALSC Models}
In this section we compare our proposed models performance with other competitive ALSC models:
\begin{itemize}
    \item \textbf{LSA+DeBERTa-V3-Large} \citep{DBLP:journals/corr/abs-2110-08604} This work introduces sentiment pattern based sentiment dependency learning framework to model the sentiment dependency between the adjacent aspects to learn the aspect polarity of an aspect without explicit sentiment context. 
    
    \item \textbf{LCF-ATEPC} \citep{DBLP:journals/ijon/YangZYSX21} This work focuses on multi-task learning based aspect term extraction and aspect sentiment polarity classification by introducing local context focusing mechanism. 
    
    \item \textbf{ABSA-DeBERTa} \citep{silvaaspect} This work introduces DeBERTa model (Decoding-enhanced BERT with Disentangled Attention) in the ALSC task and achieves competitive performance. 
    
    \item \textbf{RoBERTa+MLP} \citep{DBLP:conf/naacl/DaiYSLQ21} This work examines the effectiveness of induced tree from the Pre-trained language model and utilizes induced tree from fine-tuned RoBERTa on ALSC task to acheieve competitive performance. 
    
    \item \textbf{KaGRMN-DSG} \citep{DBLP:journals/corr/abs-2108-02352} This work utilizes both local and global syntactic representation of aspects combined with knowledge representation of aspects to enhance the overall aspect representation.  
\end{itemize}

We compare our proposed models performance with other competitive ALSC models in Table \ref{Table:10}. Our models stands 2nd and 3rd for the REST and LAPTOP domain respectively in the recent leader board for ALSC.

\subsection{Additional Results}
The extended version of Table \ref{Table:3} is given in Table \ref{Table:11} where we report the performance of our proposed ALSC models (without end-to-end training). 
Table \ref{Table:12} illustrates anecdotal examples of multi-modal aspect extraction on the three datasets for different mode importance. 



\end{document}